\documentclass{article}

\usepackage{microtype}
\usepackage{graphicx}
\usepackage{subfigure}
\usepackage{booktabs} 
\usepackage{multirow}
\usepackage{float}
\usepackage{subcaption}

\usepackage{}

\usepackage{hyperref}

\usepackage[accepted]{icml2024}

\usepackage{amsmath}
\usepackage{amssymb}
\usepackage{mathtools}
\usepackage{amsthm}

\usepackage[capitalize,noabbrev]{cleveref}

\theoremstyle{plain}

\theoremstyle{definition}

\theoremstyle{remark}

\usepackage[textsize=tiny]{todonotes}

\icmltitlerunning{Robustness Analysis of AI Models in Critical Energy Systems}

\begin{document}

\twocolumn[
\icmltitle{Robustness Analysis of AI Models in Critical Energy Systems}



\icmlsetsymbol{equal}{*}


\begin{icmlauthorlist}
\icmlauthor{Pantelis Dogoulis}{lu}
\icmlauthor{Matthieu Jimenez}{equal,lu}
\icmlauthor{Salah Ghamizi}{equal,list,riken}
\icmlauthor{Maxime Cordy}{lu}
\icmlauthor{Yves Le Traon}{lu}
\end{icmlauthorlist}

\icmlaffiliation{lu}{SerVal, University of Luxembourg, Luxembourg City, Luxembourg}

\icmlaffiliation{list}{Luxembourg Institute of Science and Technology, Luxembourg}

\icmlaffiliation{riken}{RIKEN Center for Advanced Intelligence Project (AIP), Japan}

\icmlcorrespondingauthor{Pantelis Dogoulis}{panteleimon.dogoulis@uni.lu}

\icmlkeywords{Machine Learning, ICML}

\vskip 0.3in
]



\printAffiliationsAndNotice{\icmlEqualContribution} 

\begin{abstract}
This paper analyzes the robustness of state-of-the-art AI-based models for power grid operations under the $N-1$ security criterion. While these models perform well in regular grid settings, our results highlight a significant loss in accuracy following the disconnection of a line.
Using graph theory-based analysis, we demonstrate the impact of node connectivity on this loss. Our findings emphasize the need for practical scenario considerations in developing AI methodologies for critical infrastructure.
\end{abstract}

\section{Introduction}
\label{intro}

The application of AI models has seen substantial growth across various industrial sectors, including manufacturing, transportation, and electricity. These models aim to replace costly traditional methods and improve computational efficiency. However, their deployment remains challenging due to unmet or unverified industrial criteria \cite{leyli2022lips}. The deployment of unreliable AI models in critical infrastructure poses significant risks, potentially leading to severe or catastrophic failures with dire economic and societal consequences.
In this paper, we study the robustness of state-of-the-art AI-based models in the context of power grids.
Operating power grids requires a continuous assessment of their state, which can be costly especially
the computation of power flow through grid lines. For Alternative Current (AC) grids, this problem is referred to as \textit{AC power flow} and is traditionally addressed using the Newton-Raphson method, a classical numerical approach \cite{sereeter2019comparison}. 
This method yields results with negligible error but is rather slow and lacks flexibility with regard to grid evolution.

The AC power flow problem can be characterized by a system of nonlinear equations depending on the network configuration at each time point. The network configuration encompasses the loads, generations, and topology of the network, as well as, some intrinsic characteristics of the lines (i.e. reactance, resistance). Highly trained engineers (i.e. dispatchers) can simulate the state of the power grid using these methodologies to ensure operational security, maintaining current flow within specified thresholds. 

Yet, as grid size increases, the computational complexity of this problem escalates. Adding to this, modern grids incorporate a growing amount of renewable energy sources, such as wind and solar, whose profiles are highly variable and dependent on
factors like weather, leading to significant uncertainties and further complicating the problem \cite{aslam2021survey, marot2020towards, donnot2017introducing}.

\begin{figure}[h]
  \centering
  \includegraphics[width=0.33\textwidth]{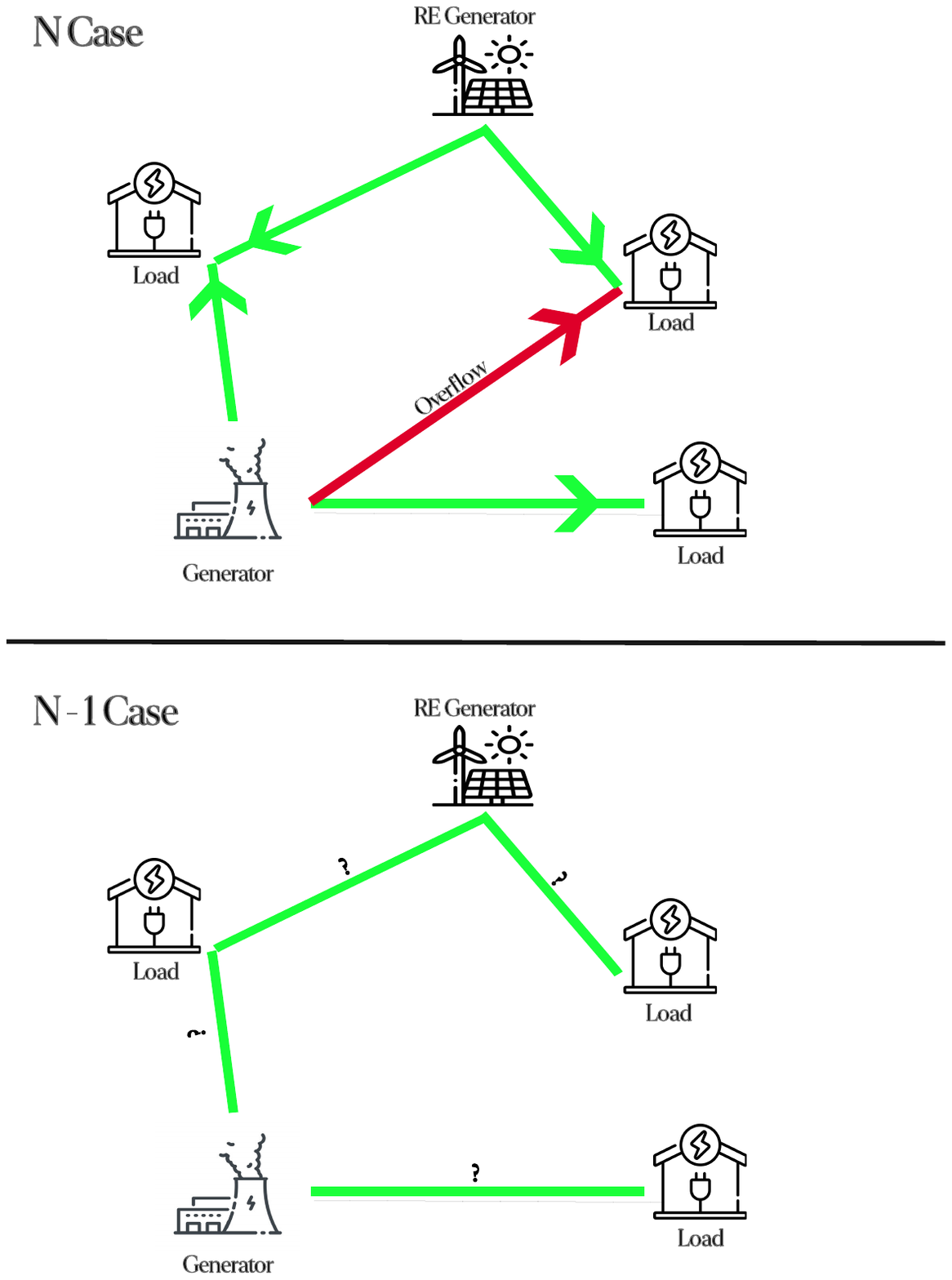}
  \caption{Example of a toy power grid featuring two generators and three loads. In the upper section of the image, an overflow in the transmission lines is depicted. To address this issue, the overloaded line is disconnected from the grid causing a topological change and necessitating a recalculation of the grid's state.}
  \label{fig:1}
\end{figure}

Therefore, researchers have naturally started to investigate AI-based approaches to address the complexity of power flow prediction problems.  This effort has been further motivated by the rise of Physics-Informed Machine Learning \cite{karniadakis2021physics}, which enables a finer integration of physical laws involved in power flow within the models \cite{hu2020physics, pagnier2021physics}. 
Nevertheless, these studies lack focus on practical scenarios, resulting in inadequate frameworks for deploying these methods. Consequently, AI methods cannot yet be deemed industry-ready, as they have been minimally tested in industrial settings and are not suitable for real-time applications.


In the context of power grids, one critical assessment is the $N-1$ security criterion. This criterion is a fundamental reliability standard used in the operation of power grids. It ensures that the grid can withstand the failure of any single transmission line, without causing widespread outages or instability. Under this criterion, the power grid is expected to continue operating even after the loss of one of its lines (Figure \ref{fig:1}). This necessitates sufficient redundancy and flexibility within the system to handle such contingencies. $N-1$ is crucial for maintaining the stability and reliability of power grids, as it helps prevent cascading failures and large-scale blackouts. Evaluating AI-based methods against it is thus essential to ensure their applicability and robustness in real-world scenarios. 
In real-time operating grids, the number of possible topological configurations increases significantly, making it computationally intensive to evaluate every potential state of the grid. While the Newton-Raphson method achieves negligible error in the $N-1$ scenario ($\approx 0$), it becomes computationally inefficient as grid size increases, due to significantly longer computation times. Conversely, machine learning models provide rapid and reliable predictions for grids without any topological modification, leveraging only inference time and achieving speeds approximately 145 times faster on large grids \cite{lin2023powerflownet}. However, in the area of grid digitization, these models must also demonstrate robustness to topological modifications within the grid to ensure its security and reliable operation.\\
In this paper, we first demonstrate the safety limitation of recent AI-based approaches for powerflow computation and then conduct an error analysis based on node connectivity.

\section{Background
}
\label{background}


\subsection{Problem Description}
 First, we can think of the grid as a graph. The nodes represent the grid's \textit{buses}, which are connected through edges, i.e. the transmission lines. The buses are categorized into three primary types: \textit{PV}, \textit{PQ}, and \textit{slack} bus. The \textit{PV} buses represent grid generators that produce and inject energy, including renewable energy (RE) generators that introduce significant uncertainties into power grid control. The \textit{PQ} buses represent the grid's loads, which are modules that consume energy. The slack bus serves as a reference point for the grid operation, where the voltage angle $\theta$ is known.
 
 The objective of the power flow prediction problem is to determine the current flowing through each transmission line based on inputs from the buses and the grid's topology. Specifically, for generator buses, the active power ($P_g$) and voltage magnitude ($V_m$) are provided, whereas, for load buses, the active ($P_l$) and reactive ($Q_l$) powers are known. 
Mathematically, the problem can be expressed through the Kirchhoff's equations.\\
\begin{equation}
\label{eq1}
\resizebox{0.43\textwidth}{!}{$
    \begin{cases}
        P_i = V_i \sum_{k=1}^{n} V_k \left( G_{ik} \cos(\theta_i - \theta_k) + B_{ik} \sin(\theta_i - \theta_k) \right) \\
        Q_i = V_i \sum_{k=1}^{n} V_k \left( G_{ik} \sin(\theta_i - \theta_k) - B_{ik} \cos(\theta_i - \theta_k) \right)
    \end{cases}
$}
\end{equation}
In the above equations, the index denotes the buses of the grid and does not follow the aforementioned notation, where the index denotes the type of the bus. However the variables remain the same, so for example $P_i$ denotes the active power of the $i$-th bus regardless of its type (i.e. generator, load, slack). Moreover $G_{ik}$ and $B_{ik}$ denote two physical line properties, the conductance and susceptance respectively. Based on this system of equations, the current that flows from bus $i$ to bus $k$ is then calculated through a basic physic's equation (Ohm's Law):
\begin{equation}
\label{eq2}
        I_{ik} = Y_{ik} (V_i - V_k)
\end{equation}
while the current injected in bus $i$, is calculated based on:
\begin{equation}
\label{eq3}
    I_i = \sum_{k=1}^{n} Y_{ik} V_k    
\end{equation}
In Eqs. \ref{eq2} and \ref{eq3}, $Y_{ik}$ is the element in the $(i,k)$ position of the admittance matrix $Y$. This matrix consists of the admittance values (i.e. physical characteristics of the transmission lines) between different nodes in the power grid.

\paragraph{Newton-Raphson method}
Traditionally, AC power flow problem solvers use the Newton-Raphson method, linearizing power flow equations around an initial guess, $\mathbf{x}^{0}$. The equations are represented as:
\begin{equation}
    \mathbf{F}(\mathbf{x}) = 
    \begin{bmatrix}
        P_i - P_i^{\text{appr}}(\mathbf{x}) \\
        Q_i - Q_i^{\text{appr}}(\mathbf{x})
    \end{bmatrix} 
    = \mathbf{0},
\end{equation}

where $P_i$ and $Q_i$ are the specified active and reactive powers for each bus $i$, and $P_i^{\text{appr}}$ and $Q_i^{\text{appr}}$ are the calculated powers based on the state vector $\mathbf{x}$ (voltage magnitudes and angles). The iterative update is given by:
\begin{equation}
    \mathbf{x}^{(k+1)} = \mathbf{x}^{(k)} - \mathbf{J}^{-1}(\mathbf{x}^{(k)}) \mathbf{F}(\mathbf{x}^{(k)}),
\end{equation}
where $\mathbf{J}$ is the Jacobian matrix of partial derivatives of $\mathbf{F}$ with respect to $\mathbf{x}$. This process is repeated until reaching the convergence threshold defined by the user.\\
Unlike traditional numerical approaches, machine learning methods aim to predict current flows based on bus inputs and grid topology. These methods estimate the function $f$ that:
\begin{equation}
    f: (X, \tau) \rightarrow I
\end{equation}
where $X$ denotes the space of the inputs of the buses, $\tau$ denotes the space of the grid's topology and $I$ denotes the the output space. \\
The dimension of $I$ varies across studies and depends on the framework employed.  Some research predicts the vector $(P_i$, $Q_i$, $V_i$, $\theta_i)$ for each bus, while others predict directly the vector $(I_i, I_{ik})$. Although using the Eqs. \ref{eq2} and \ref{eq3}, one can infer the former vector from the latter and vice versa.

\subsection{Related Work}
In recent years, the demand for faster and more robust power grid simulators has increased due to the growing complexity of grid control problems and traditional simulation approaches \cite{sereeter2019comparison, kulworawanichpong2010simplified, d2021comparing, coffrin2014linear, capitanescu2016critical} have started to open the path for AI-based ones to meet rising computational requirements and the integration of renewable energy. 
One of the primary approaches was introduced by \cite{donon2020neural} where the authors suggested a graph neural network architecture capable of calculating power flow and generalizing to small and medium-sized grids. The originality exists in the direct incorporation of Kirchhoff's law into the optimization objective. 
Similarly, \cite{lin2023powerflownet} introduced a graph-based architecture that also integrates Kirchhoff's law into the loss function and employs a message-passing mechanism for feature propagation during training. Other relevant approaches include the work of \cite{bolz2019power,ghamizi2024powerflowmultinet}, who suggested a graph-based convolutional neural network. \cite{donnot2018fast} also proposed a novel dropout mechanism \cite{srivastava2014dropout} within a simple feed-forward neural network to train the model using $N-1$ instances. Building on this approach, \cite{donon2020leap} proposed an architecture inspired by the aforementioned dropout technique. Their model incorporates an additional block designed to handle topological changes, and the training process includes both instances with the original topology ($N$ case) and $N-1$ cases, enhancing the model's adaptability and robustness.

\section{Experimental Setup}
\label{setup}
 Our aim is to evaluate the recently suggested AI-based approaches for power flow prediction on the $N-1$ security criterion, thus assessing their industrial readiness. In particular, we perform evaluations over two datasets of $N-1$ cases and explore whether: \textit{(a)} the degree of connectivity of a node for which a line has been disconnected impacts the robustness of the approach to $N-1$, \textit{(b)} prior exposure of the model to certain $N-1$ instances enhances its robustness.

\subsection{Datasets}
To perform our experiments, we rely on two standard datasets across the Power Engineering community, namely IEEE 14 and IEEE 118. These datasets contain one instance of the grid, with the nominal values for each bus. They consist of 14 and 118 busses respectively.\\
To approach the problem from a machine learning perspective, it is required to augment these datasets to generate new \textit{iid} data instances. For this purpose, we employed two distinct strategies: \textit{(a)} Constrained Sampling and \textit{(b)} Random Agent Cutting. 

Regarding \textit{(a)}, we sampled new data instances for PQ buses based on a normal distribution around their nominal values while for the generators based on Dirichlet simplex sampling. Mathematically we express the process as: \\$P_l \sim N(|P_l|, 0.01)$, $V_m \sim N(|V_m|, 0.01)$,\\ $Q_l \sim N(|Q_l|, 0.01)$ and $P_g \sim Dir(G)$,\\ where $G$ denotes the total amount of generation.

We use Dirichlet sampling for the generators, since the total amount of generation should be constant among all the data instances, and is predefined in the grid's configuration. Then, we utilize Newton-Raphson method to calculate the outputs of the power flow problem (ground truth). The output is a vector $(P_i, Q_i, V_i, \theta_i)$ for each bus $i$ of the generated power grid instance. 

For the second strategy \textit{(b)}, a grid agent was initialized and tasked with randomly cutting one line of the grid, based on a probability $p$. The new state of the grid was computed using the Newton-Raphson method (same as before), and the resulting data instance was recorded.

\subsection{Models}
We chose 3 state-of-the-art models namely: PowerFlowNet \cite{lin2023powerflownet}, LeapNet \cite{donon2020leap} and ResNet \cite{donon2020leap}. The models relies on the same inputs $X$ which are the known values of each bus, but differ in their outputs differ. Hence, PowerFlowNet  outputs the vector $(P_i, Q_i, V_i, \theta_i)$ for each bus $i$, while LeapNet and Resnet outputs the vector $(I_i, I_{ik})$. Both vectors can be converted into the other using Eqs. \ref{eq2} and \ref{eq3}.

\subsection{Evaluation Protocol}
Firstly, we utilized the pretrained PowerFlowNet models in the aforementioned datasets and we evaluated them in the $N-1$ cases. We note that these models were originally trained exclusively on $N$ cases, meaning they had no exposure to topological changes in the grid during training. We also trained a modified version of ResNet\footnote{This is not the classical version of ResNet as defined in \cite{he2016deep}.  Instead, we adopt the implementation detailed in \cite{donon2020leap}, where the authors refer to their model as ResNet, inspired by the incorporation of residual connections.}, with the final linear layer replaced with one that projects the data into $\mathbb{R}^2$ to match the dimensions of the output \textit{I}. We then applied the $N-1$ criterion to assess the robustness of these models in real and critical scenarios. Moreover, training with instances of $N-1$ was performed in order to enhance the robustness of the models. For this, we also used LeapNet which integrates a module into its architecture that tracks the topological changes of the grid.

\subsection{Metric}

To evaluate the performance of the models, we used the traditional Mean Squared Error (MSE) metric. For the ResNet and LeapNet models, performance was assessed by averaging the MSE of the outputs $I_i$ and $I_{ik}$, representing the current at both the origin and end of the line. Regarding PowerFlowNet as the model outputs a different vector, we use the equations to convert the output into $I_{i}$ and $I_{ik}$ to obtain comparable result.

\subsection{Implementation Details}

We explore 3 configurations of the PowerFlowNet, defined as $PowerFlowNet_L$ (large), $PowerFlowNet_M$ (medium), and $PowerFlowNet_S$ (small). These reflect the number of hidden topology-adaptive convolutions utilized, as specified in the original paper. 
In a similar manner, we implemented $LeapNet_M$, $LeapNet_S$, $ResNet_M$, and $ResNet_S$ which vary in the number of hidden layers used, according to the values of \cite{donon2020leap}. 
\\
For the dataset generation we used Grid2op \cite{grid2op} and Pandapower \cite{thurner2018pandapower}, which are two popular Python frameworks for AI-based Power Engineering. The models were trained for 25 epochs using the Adam optimizer \cite{kingma2014adam}.  The learning rate was fixed at $0.001$, with a linear scheduler of a step size equal to 5 and a batch size equal to 128. The training dataset comprised 10,000 different $N$ grid instances, while both the evaluation dataset for the $N$ case and the evaluation dataset for the $N-1$ case consisted of 2,000 instances each.

\begin{figure*}[h!]
    \centering
    \subfigure[Normal]{\label{a}\includegraphics[width=42mm]{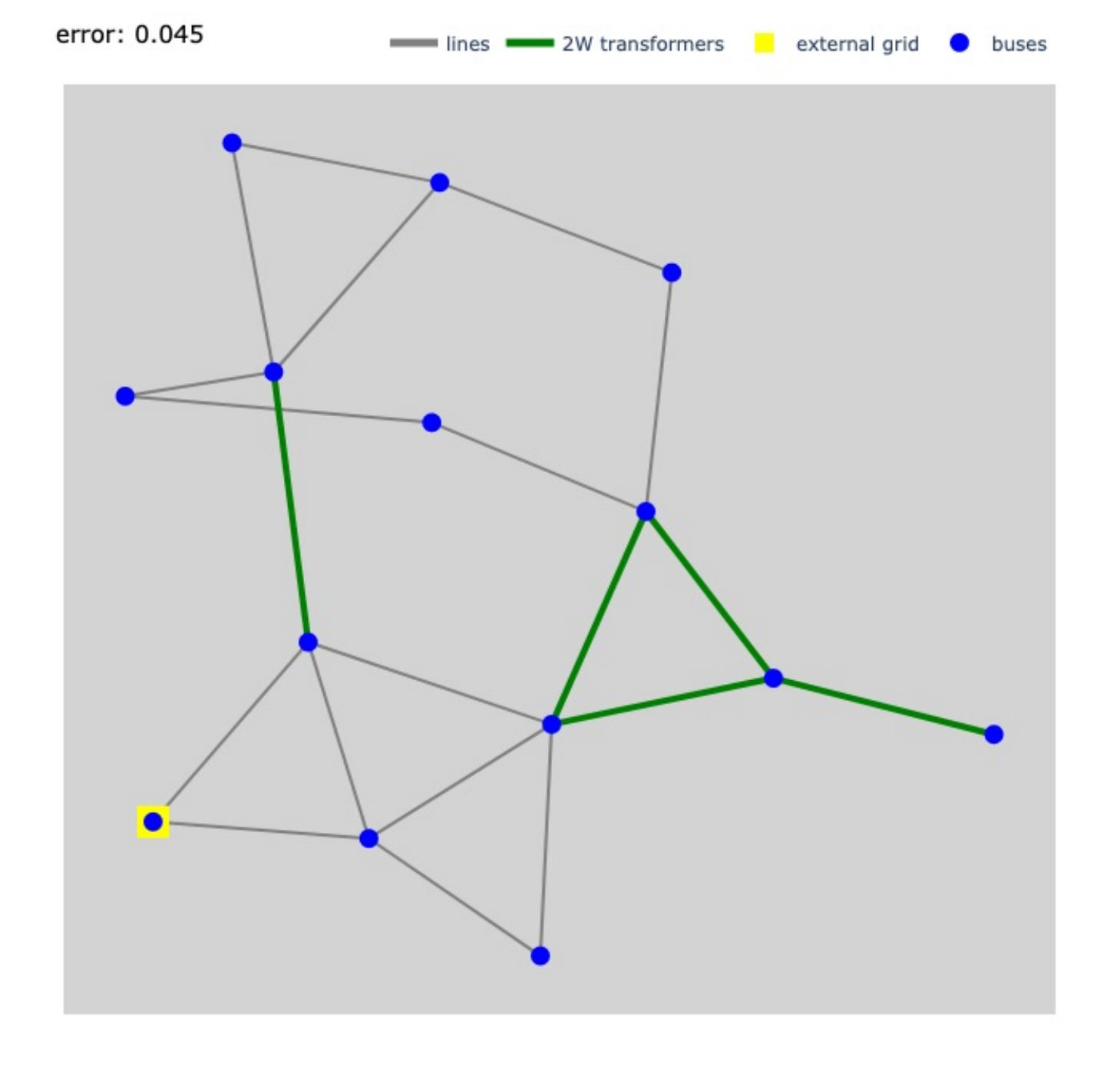}}
    \subfigure[1 $\rightarrow$ 2]{\label{b}\includegraphics[width=42mm]{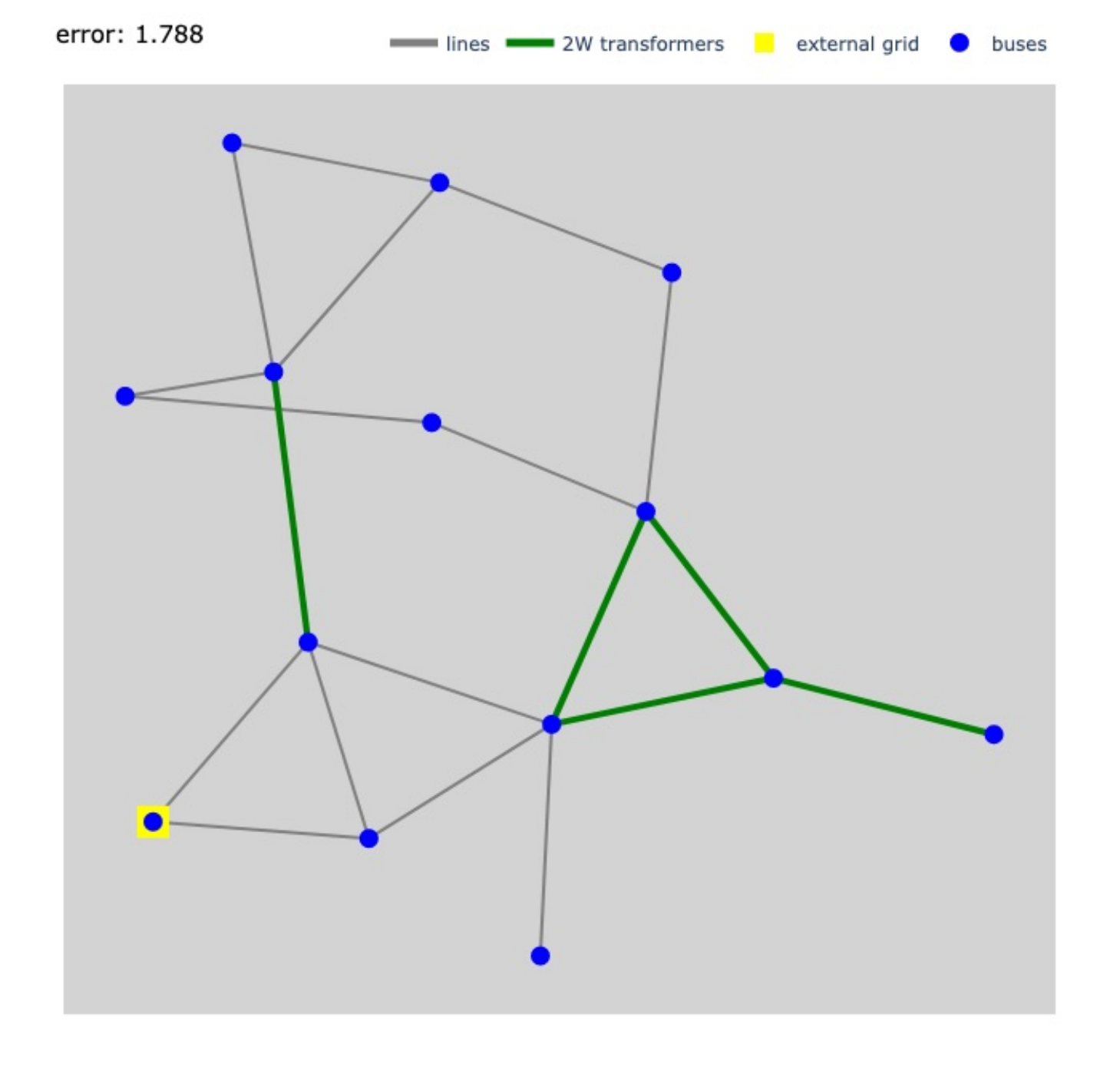}}
    \subfigure[1 $\rightarrow$ 3]{\label{c}\includegraphics[width=42mm]{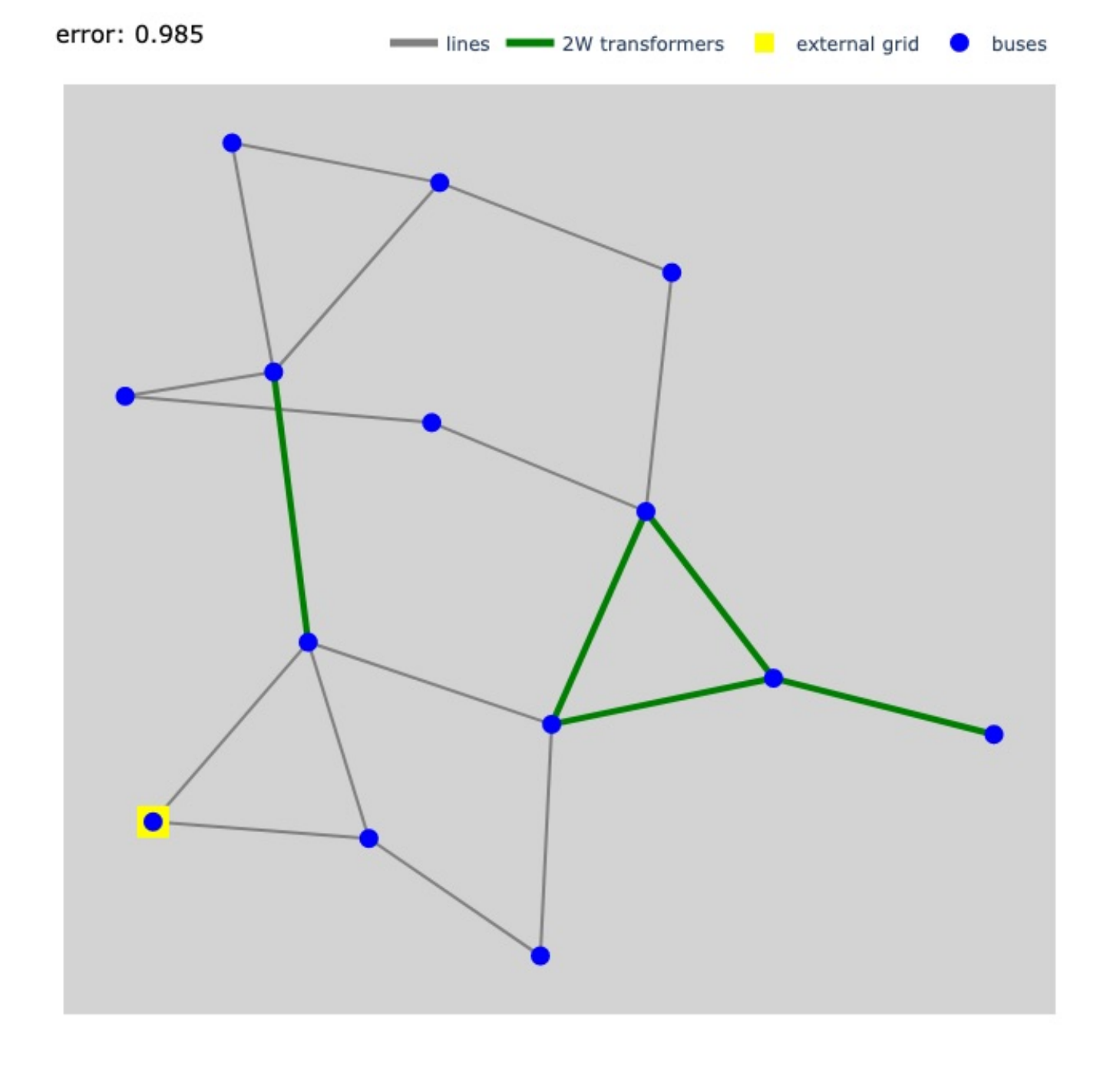}}
    \subfigure[1 $\rightarrow$ 4]{\label{d}\includegraphics[width=42mm]{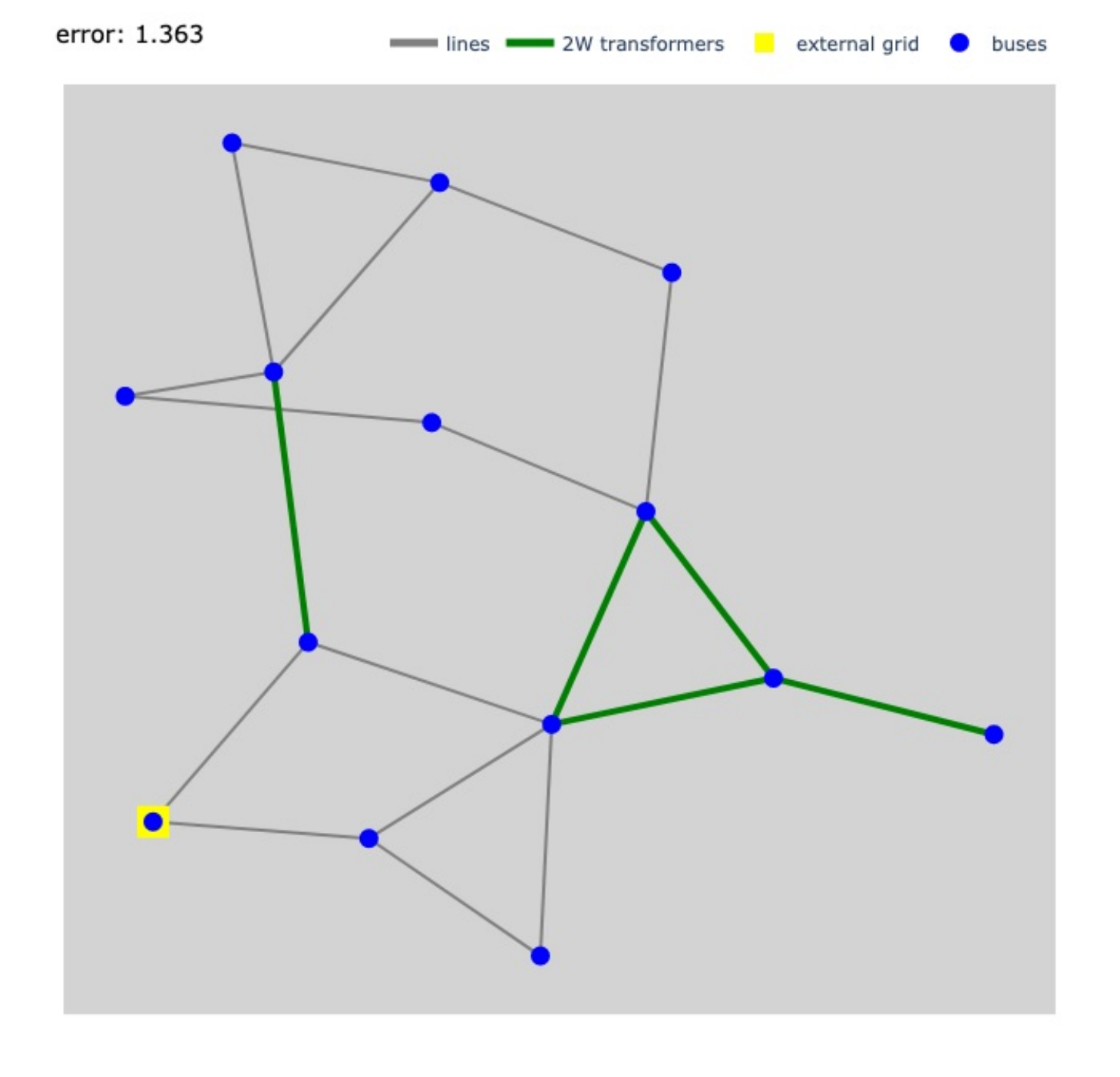}}
    \caption{Visual comparison of the model's prediction in the $N$ case and three different $N-1$ cases for the IEEE 14 dataset. In (\textbf{a}), the standard topology is shown. In (\textbf{b}), the line from bus 1 to bus 2 is disconnected. In (\textbf{c}), the line from bus 1 to bus 3 is disconnected. In (\textbf{d}), the line from bus 1 to bus 4 is disconnected.}
    \label{fig:3}
\end{figure*}

\vspace{-0.5em}
\section{Results}
\label{results}
In this section, we present the experimental results. Initially, we demonstrate that pretrained models lack robustness in the $N-1$ scenario and we conduct a short analysis of this issue using graph theory. Finally, as a partial solution to this problem, we report on the outcomes of an empirical study by mixing $N-1$ cases into the training set.
\vspace{-0.5em}
\subsection{$N-1$ case}
While the results on the $N$ cases are promising for each approach on the two datasets, and could even be deemed reliable enough given the speedup compared to Newton Raphson, their performance deteriorates in the modified topology scenario. We observe a significant increase in the error when evaluating the models under the $N-1$ scenario (Table \ref{tab:1}). 
In particular, the best MSE reported for the $N$ case is $0.051$, resp. $0.113$, whereas for the $N-1$ case it is $1.467$, resp. $4.184$ in the IEEE 14 dataset, resp. in the IEEE 118 dataset. The increase ranges between 10-100x in most cases and highlights the extent to which the models are impacted by simple grid's topology modification, especially in the case of larger grids such as IEEE 118, which remains small in comparison to actual grids.\\ 

This indicates that the current models lack robustness when subjected to the critical assessments required for real-time operating grids.

\begin{table}[h!]
    \centering
    \begin{tabular}{cccc}
        \toprule
        \textbf{Models} & \textbf{Datasets} & \textbf{N} & \textbf{N-1} \\
        \midrule
        \multirow{2}{*}{$PowerFlowNet_L$} & IEEE 14 & $\textbf{0.051}$ & $\textbf{1.467}$ \\
                                          & IEEE 118 & $0.363$ & $8.293$ \\
        \midrule
        \multirow{2}{*}{$PowerFlowNet_M$} & IEEE 14 & $0.076$ & $1.582$ \\
                                          & IEEE 118 & $0.184$ & $\color{gray}\textbf{4.184}$ \\
        \midrule
        \multirow{2}{*}{$PowerFlowNet_S$} & IEEE 14 & $0.055$ & $2.358$ \\
                                          & IEEE 118 & $0.415$ & $7.249$ \\
        \midrule
        \multirow{2}{*}{$ResNet_M$} & IEEE 14 & $0.089$ & $1.789$ \\
                                     & IEEE 118 & $0.594$ & $4.536$ \\
        \midrule
        \multirow{2}{*}{$ResNet_S$} & IEEE 14 & $0.097$ & $2.121$ \\
                                     & IEEE 118 & $\color{gray}\textbf{0.113}$ & $7.897$ \\
        \bottomrule
    \end{tabular}
    \caption{MSE for the power flow prediction without any topological change and for the $N-1$ case, across the two datasets. \textbf{Bold} indicates the smaller MSE reported for the IEEE 14 dataset, while {\color{gray}\textbf{gray}} indicates the smaller MSE reported for the IEEE 118 dataset.}
    \label{tab:1}
\end{table}

\subsection{Graph Analysis}
\label{graph}

\begin{table}[h!]
    \centering
    \begin{tabular}{ccccc}
        \toprule
        \textbf{Dataset} & $\mathcal{D}_{max}$ & \textbf{MSE} & $\mathcal{D}_{median}$ & \textbf{MSE} \\
        \midrule
        \multirow{2}{*}{IEEE 14} & 5 & \textbf{1.297} & 3 & 1.325 \\
                                                 & 4 & 1.632 & 2 & \textbf{1.244} \\
        \midrule
        \multirow{2}{*}{IEEE 118} & 12 & \textbf{7.526} & 8 & \textbf{4.864} \\
                                                 & 11 & 8.491 & 7 & 6.503 \\
        \bottomrule
    \end{tabular}
    \caption{Comparison of the MSE metric based on the node with the highest degree $\mathcal{D}_{max}$ and on the node with the median degree $\mathcal{D}_{median}$ for the IEEE 14 and IEEE 118 datasets. \textbf{Bold} indicates the smaller MSE reported.}
    \label{tab:2}
\end{table}

To further investigate the issue, we conducted an additional analysis to determine if the model's performance in the $N-1$ scenario is influenced by the specific line removed from the grid topology. \\
We, therefore, focused on nodes with higher connectivity to assess their impact on model predictions (Figure \ref{fig:3}). Specifically, we identified the node with the highest degree $x_{\mathcal{D}_{max}}$ in the original topology ($N$ case). Since the grid agent removes lines randomly, the degree of $x_{\mathcal{D}_{max}}$ varies across instances. We clustered test set instances based on the degree of the node $x_{\mathcal{D}_{max}}$ and evaluated the model on these subsets. Additionally, we included the node with the median degree $x_{\mathcal{D}_{median}}$ for a more comprehensive analysis.
For this experiment, we utilized $PowerFlowNet_L$, which demonstrated the lowest MSE error in both the $N$ and $N-1$ power flow scenarios for the IEEE 14 dataset.

In Table \ref{tab:2}, for each dataset, the first row represents the original values of $\mathcal{D}{max}$ and $\mathcal{D}{median}$, while the second row shows the degrees of these nodes when a connected line is removed. We observe that when a line connected to nodes with higher degrees is removed, it has a greater impact on the prediction process, as it produces a bigger error in the model. The intuition behind this finding is that nodes with higher degrees exhibit greater connectivity within the graph, thereby exerting a more significant impact on the grid's physical behavior.
\begin{table}[h!]
    \centering
    \begin{tabular}{ccc}
        \toprule
        \textbf{Models} & \textbf{p} & \textbf{MSE} \\
        \midrule
        \multirow{2}{*}{$PowerFlowNet_L$} & $0.01$ & $\textbf{0.141}$ \\
                                                 & $0.1$ & $0.078$ \\
        \midrule
        \multirow{2}{*}{$PowerFlowNet_M$} & $0.01$ & $0.358$ \\
                                                 & $0.1$ & $0.091$ \\
        \midrule
        \multirow{2}{*}{$PowerFlowNet_S$} & $0.01$ & $0.195$ \\
                                                 & $0.1$ & $0.096$ \\
        \midrule
        \multirow{2}{*}{$ResNet_M$} & $0.01$ & $0.296$ \\
                                           & $0.1$ & $0.104$ \\
        \midrule
        \multirow{2}{*}{$ResNet_S$} & $0.01$ & $0.348$ \\
                                            & $0.1$ & $0.122$ \\
        \midrule
        \multirow{2}{*}{$LeapNet_M$} & $0.01$ & $0.175$ \\
                                            & $0.1$ & $\textbf{\color{gray}0.067}$ \\
        \midrule
        \multirow{2}{*}{$LeapNet_S$} & $0.01$ & $0.137$ \\
                                           & $0.1$ & $0.082$ \\

        \bottomrule
    \end{tabular}
    \caption{MSE for the mixed $N-1$ and \textit{N-2} power flow prediction across the two datasets. \textbf{Bold} indicates the smaller MSE reported for the $p=0.01$, while {\color{gray}\textbf{gray}} indicates the smaller MSE reported for $p=0.1$.}
    \label{tab:3}
\end{table}

\subsection{Mix Training}

The ResNet, LeapNet, and PowerFlowNet models were trained using instances of both $N$ and $N-1$ cases from the IEEE 14 dataset to evaluate their performance on unseen $N-1$ cases. The probability $p$ of generating an $N-1$ instance was set to either $0.01$ or $0.1$ for each dataset. As shown in Table \ref{tab:3}, the robustness of each model notably improves, particularly when $p=0.1$. This finding is particularly significant for larger grids, where the number of $N-1$ configurations increases exponentially. It demonstrates that grid robustness is enhanced even with a relatively small number of $N-1$ instances during the training process.

\section{Conclusion}
\label{conclusion}

In this paper, we analyzed the robustness of AI-based models for power grid operations under the $N-1$ security criterion. Our study identified significant deficiencies in model robustness under real-time conditions and emphasized the need for improvements. A graph theory-based analysis highlighted the critical impact of node connectivity on model predictions. We deployed a mixed training technique with both $N$ and $N-1$ instances, which significantly enhanced model adaptability and reliability. Future research could focus on developing a sampling method for $N-1$ instances to construct a training dataset, taking into account graph connectivity. This approach is motivated by the observation that the connectivity degree of each node significantly impacts the prediction accuracy of the models. In conclusion, our findings stress the importance of practical scenarios in developing AI methodologies for critical infrastructure, advocating for future research on more complex topological changes and related topics, such as Voltage Control \cite{sun2019review}.

\bibliography{grid_rob}

\begin{thebibliography}{24}
\providecommand{\natexlab}[1]{#1}
\providecommand{\url}[1]{\texttt{#1}}
\expandafter\ifx\csname urlstyle\endcsname\relax
  \providecommand{\doi}[1]{doi: #1}\else
  \providecommand{\doi}{doi: \begingroup \urlstyle{rm}\Url}\fi

\bibitem[Aslam et~al.(2021)Aslam, Herodotou, Mohsin, Javaid, Ashraf, and Aslam]{aslam2021survey}
Aslam, S., Herodotou, H., Mohsin, S.~M., Javaid, N., Ashraf, N., and Aslam, S.
\newblock A survey on deep learning methods for power load and renewable energy forecasting in smart microgrids.
\newblock \emph{Renewable and Sustainable Energy Reviews}, 144:\penalty0 110992, 2021.

\bibitem[Bolz et~al.(2019)Bolz, Rue{\ss}, and Zell]{bolz2019power}
Bolz, V., Rue{\ss}, J., and Zell, A.
\newblock Power flow approximation based on graph convolutional networks.
\newblock In \emph{2019 18th ieee international conference on machine learning and applications (icmla)}, pp.\  1679--1686. IEEE, 2019.

\bibitem[Capitanescu(2016)]{capitanescu2016critical}
Capitanescu, F.
\newblock Critical review of recent advances and further developments needed in ac optimal power flow.
\newblock \emph{Electric Power Systems Research}, 136:\penalty0 57--68, 2016.

\bibitem[Coffrin \& Van~Hentenryck(2014)Coffrin and Van~Hentenryck]{coffrin2014linear}
Coffrin, C. and Van~Hentenryck, P.
\newblock A linear-programming approximation of ac power flows.
\newblock \emph{INFORMS Journal on Computing}, 26\penalty0 (4):\penalty0 718--734, 2014.

\bibitem[Donnot(2020)]{grid2op}
Donnot, B.
\newblock {Grid2op- A testbed platform to model sequential decision making in power systems. }.
\newblock \url{https://GitHub.com/rte-france/grid2op}, 2020.

\bibitem[Donnot et~al.(2017)Donnot, Guyon, Schoenauer, Panciatici, and Marot]{donnot2017introducing}
Donnot, B., Guyon, I., Schoenauer, M., Panciatici, P., and Marot, A.
\newblock Introducing machine learning for power system operation support.
\newblock \emph{arXiv preprint arXiv:1709.09527}, 2017.

\bibitem[Donnot et~al.(2018)Donnot, Guyon, Schoenauer, Marot, and Panciatici]{donnot2018fast}
Donnot, B., Guyon, I., Schoenauer, M., Marot, A., and Panciatici, P.
\newblock Fast power system security analysis with guided dropout.
\newblock \emph{arXiv preprint arXiv:1801.09870}, 2018.

\bibitem[Donon et~al.(2020{\natexlab{a}})Donon, Cl{\'e}ment, Donnot, Marot, Guyon, and Schoenauer]{donon2020neural}
Donon, B., Cl{\'e}ment, R., Donnot, B., Marot, A., Guyon, I., and Schoenauer, M.
\newblock Neural networks for power flow: Graph neural solver.
\newblock \emph{Electric Power Systems Research}, 189:\penalty0 106547, 2020{\natexlab{a}}.

\bibitem[Donon et~al.(2020{\natexlab{b}})Donon, Donnot, Guyon, Liu, Marot, Panciatici, and Schoenauer]{donon2020leap}
Donon, B., Donnot, B., Guyon, I., Liu, Z., Marot, A., Panciatici, P., and Schoenauer, M.
\newblock Leap nets for system identification and application to power systems.
\newblock \emph{Neurocomputing}, 416:\penalty0 316--327, 2020{\natexlab{b}}.

\bibitem[D’orto et~al.(2021)D’orto, Sj{\"o}blom, Chien, Axner, and Gong]{d2021comparing}
D’orto, M., Sj{\"o}blom, S., Chien, L.~S., Axner, L., and Gong, J.
\newblock Comparing different approaches for solving large scale power-flow problems with the newton-raphson method.
\newblock \emph{IEEE Access}, 9:\penalty0 56604--56615, 2021.

\bibitem[Ghamizi et~al.(2024)Ghamizi, Cao, Ma, and Rodriguez]{ghamizi2024powerflowmultinet}
Ghamizi, S., Cao, J., Ma, A., and Rodriguez, P.
\newblock Powerflowmultinet: Multigraph neural networks for unbalanced three-phase distribution systems.
\newblock \emph{arXiv preprint arXiv:2403.00892}, 2024.

\bibitem[He et~al.(2016)He, Zhang, Ren, and Sun]{he2016deep}
He, K., Zhang, X., Ren, S., and Sun, J.
\newblock Deep residual learning for image recognition.
\newblock In \emph{Proceedings of the IEEE conference on computer vision and pattern recognition}, pp.\  770--778, 2016.

\bibitem[Hu et~al.(2020)Hu, Hu, Verma, and Zhang]{hu2020physics}
Hu, X., Hu, H., Verma, S., and Zhang, Z.-L.
\newblock Physics-guided deep neural networks for power flow analysis.
\newblock \emph{IEEE Transactions on Power Systems}, 36\penalty0 (3):\penalty0 2082--2092, 2020.

\bibitem[Karniadakis et~al.(2021)Karniadakis, Kevrekidis, Lu, Perdikaris, Wang, and Yang]{karniadakis2021physics}
Karniadakis, G.~E., Kevrekidis, I.~G., Lu, L., Perdikaris, P., Wang, S., and Yang, L.
\newblock Physics-informed machine learning.
\newblock \emph{Nature Reviews Physics}, 3\penalty0 (6):\penalty0 422--440, 2021.

\bibitem[Kingma \& Ba(2014)Kingma and Ba]{kingma2014adam}
Kingma, D.~P. and Ba, J.
\newblock Adam: A method for stochastic optimization.
\newblock \emph{arXiv preprint arXiv:1412.6980}, 2014.

\bibitem[Kulworawanichpong(2010)]{kulworawanichpong2010simplified}
Kulworawanichpong, T.
\newblock Simplified newton--raphson power-flow solution method.
\newblock \emph{International journal of electrical power \& energy systems}, 32\penalty0 (6):\penalty0 551--558, 2010.

\bibitem[Leyli~Abadi et~al.(2022)Leyli~Abadi, Marot, Picault, Danan, Yagoubi, Donnot, Attoui, Dimitrov, Farjallah, and Etienam]{leyli2022lips}
Leyli~Abadi, M., Marot, A., Picault, J., Danan, D., Yagoubi, M., Donnot, B., Attoui, S., Dimitrov, P., Farjallah, A., and Etienam, C.
\newblock Lips-learning industrial physical simulation benchmark suite.
\newblock \emph{Advances in Neural Information Processing Systems}, 35:\penalty0 28095--28109, 2022.

\bibitem[Lin et~al.(2023)Lin, Orfanoudakis, Cardenas, Giraldo, and Vergara]{lin2023powerflownet}
Lin, N., Orfanoudakis, S., Cardenas, N.~O., Giraldo, J.~S., and Vergara, P.~P.
\newblock Powerflownet: Leveraging message passing gnns for improved power flow approximation.
\newblock \emph{arXiv preprint arXiv:2311.03415}, 2023.

\bibitem[Marot et~al.(2020)Marot, Rozier, Dussartre, Crochepierre, and Donnot]{marot2020towards}
Marot, A., Rozier, A., Dussartre, M., Crochepierre, L., and Donnot, B.
\newblock Towards an ai assistant for power grid operators.
\newblock \emph{arXiv preprint arXiv:2012.02026}, 2020.

\bibitem[Pagnier \& Chertkov(2021)Pagnier and Chertkov]{pagnier2021physics}
Pagnier, L. and Chertkov, M.
\newblock Physics-informed graphical neural network for parameter \& state estimations in power systems.
\newblock \emph{arXiv preprint arXiv:2102.06349}, 2021.

\bibitem[Sereeter et~al.(2019)Sereeter, Vuik, and Witteveen]{sereeter2019comparison}
Sereeter, B., Vuik, C., and Witteveen, C.
\newblock On a comparison of newton--raphson solvers for power flow problems.
\newblock \emph{Journal of Computational and Applied Mathematics}, 360:\penalty0 157--169, 2019.

\bibitem[Srivastava et~al.(2014)Srivastava, Hinton, Krizhevsky, Sutskever, and Salakhutdinov]{srivastava2014dropout}
Srivastava, N., Hinton, G., Krizhevsky, A., Sutskever, I., and Salakhutdinov, R.
\newblock Dropout: a simple way to prevent neural networks from overfitting.
\newblock \emph{The journal of machine learning research}, 15\penalty0 (1):\penalty0 1929--1958, 2014.

\bibitem[Sun et~al.(2019)Sun, Guo, Qi, Ajjarapu, Bravo, Chow, Li, Moghe, Nasr-Azadani, Tamrakar, et~al.]{sun2019review}
Sun, H., Guo, Q., Qi, J., Ajjarapu, V., Bravo, R., Chow, J., Li, Z., Moghe, R., Nasr-Azadani, E., Tamrakar, U., et~al.
\newblock Review of challenges and research opportunities for voltage control in smart grids.
\newblock \emph{IEEE Transactions on Power Systems}, 34\penalty0 (4):\penalty0 2790--2801, 2019.

\bibitem[Thurner et~al.(2018)Thurner, Scheidler, Sch{\"a}fer, Menke, Dollichon, Meier, Meinecke, and Braun]{thurner2018pandapower}
Thurner, L., Scheidler, A., Sch{\"a}fer, F., Menke, J.-H., Dollichon, J., Meier, F., Meinecke, S., and Braun, M.
\newblock pandapower—an open-source python tool for convenient modeling, analysis, and optimization of electric power systems.
\newblock \emph{IEEE Transactions on Power Systems}, 33\penalty0 (6):\penalty0 6510--6521, 2018.

\end{thebibliography}
\bibliographystyle{icml2024}
\end{document}